\documentclass[twoside]{article}
\usepackage{ecj,palatino,epsfig,latexsym,natbib,amsmath,amssymb}
\usepackage[table,xcdraw]{xcolor}

\begin{document}

\ecjHeader{x}{x}{xxx-xxx}{201X}{Role of Morphological Variation in Evolutionary Robotics}{Carvalho, J. T. \& Nolfi, S.}
\title{\bf The Role of Morphological Variation in Evolutionary Robotics: Maximizing Performance and Robustness}  

\author{\name{\bf Jonata Tyska  Carvalho $^{1,2}$} \hfill \addr{jonata.tyska@ufsc.br}\\ 
        $^{1}$ \addr{Informatics and Statistics Department, Federal University of Santa Catarina (UFSC), 
        Florianópolis, Brazil}
\AND
       \name{\bf Stefano Nolfi $^{2}$} \hfill \addr{stefano.nolfi@istc.cnr.it}\\
        $^{2}$ \addr{Institute of Cognitive Sciences and Technologies (ISTC), National Research Council (CNR), 
        Rome, Italy}
}

\maketitle

\begin{abstract}

Exposing an Evolutionary Algorithm that is used to evolve robot controllers to variable conditions is necessary to obtain solutions which are robust and can cross the reality gap. However, we do not yet have methods for analyzing and understanding the impact of the varying morphological conditions which impact the evolutionary process, and therefore for choosing suitable variation ranges. By morphological conditions, we refer to the starting state of the robot, and to variations in its sensor readings during operation due to noise. In this article, we introduce a method that permits us to measure the impact of these morphological variations and we analyze the relation between the amplitude of variations, the modality with which they are introduced, and the performance and robustness of evolving agents. Our results demonstrate that (i) the evolutionary algorithm can tolerate morphological  variations which have a very high impact, (ii) variations affecting the actions of the agent are tolerated much better than variations affecting the initial state of the agent or of the environment, and (iii) improving the accuracy of the fitness measure through multiple evaluations is not always useful. Moreover, our results show that morphological variations permit generating solutions which perform better both in varying and non-varying conditions.

\end{abstract}


\section{Introduction}
Evolving robots~\citep{nolfi2016evolutionary} are often evaluated in variable initial and environmental conditions. This is necessary to select robots that are robust to environmental variations~\citep{branke2012evolutionary,bredeche2012environment,milano2019moderate,pagliuca2019robust} and which can eventually adapt to varying conditions on the fly~\citep{floreano2000evolutionary,risi2012unified,cully2015robots}. The physical world is highly uncertain, almost no characteristic, dimension, or property remains constant. Selecting individuals that are robust with respect to these forms of variations is thus essential to obtain robots capable of operating in a natural environment. Exposing evolving robots to variable conditions is also necessary to cross the reality gap, i.e., to transfer robots trained in simulation in hardware successfully~\citep{jakobi1995noise,koos2012transferability,sadeghi2016cad2rl,salvato2021}. Indeed, the ability to cope with variations can enable the robots to cope also with the differences between the simulated and real environment.

The presence of variations during individuals' evaluation makes the fitness measure stochastic. This is caused by the fact that, in the presence of variations, the fitness reflects both the ability of the agent and the relative difficulty of the conditions encountered during the evaluation of the agent. The higher the variance of variations, the higher the chances that the fitness is over or under-estimated. The range of variations should thus be set in a way that maximizes the robustness of the solutions and which ensures that the information provided by the fitness is not too low. 

In this article, we introduce a method that permits us to measure the impact of morphological variations on the fitness measure in the specific experimental setting considered. By morphological variations, we mean variations on the initial posture of the agents and in the effectiveness of their actuators, i.e., actuators that are more or less noisy. Therefore, we study the impact of the modality with which variations are introduced on the performance and robustness of the evolving robots. More specifically, we analyze the effect of variations affecting the initial state of the robot and of variations affecting the state of the robot in each step by changing the noise level present in their actuators. Finally, we analyze the effect of increasing the variance of variations during evaluation episodes or across generations. 

Overall, our results demonstrate that: (i) the evolutionary algorithm we tested can tolerate variations which have a very high impact on the fitness of the robots, (ii) variations affecting the actions of the robot during each step are tolerated much better than variations affecting the initial state of the robot, (iii) improving the precision of the fitness measure through multiple evaluations is not necessarily useful, and (iv) robots exposed to variations often outperform robots not exposed to variations also in non-varying conditions.  

Overall, the knowledge and method herein presented can enable experimenters to set suitable hyper-parameters in a principled way and to obtain better results. 

\section{Related work}

Evolutionary algorithms are intrinsically robust to variations and to the stochasticity of the fitness thanks to the fact that they operate on a population of candidate solutions ~\citep{arnold2002noisy}. Clearly, however, they can fail to produce progress when the amount of variation is very large and, consequently, when the signal-to-noise ratio is too small. 

Previous works demonstrated that the possibility to cope with variations can be increased by enlarging the size of the population \citep{arnold2002noisy,harik1999gambler}. Other works demonstrated how the possibility to cope with variations can be increased by using the average fitness collected during multiple evaluation episodes ~\citep{aizawa1994scheduling,branke2003selection,stagge1998averaging,cantu2004adaptive,hansen2008method}. Indeed, calculating the fitness on multiple evaluation episodes during which the robot is exposed to varying environmental conditions permits to reduce the stochasticity of the fitness. Clearly, both methods are computationally expensive. For this reason, several authors proposed methods that vary the number of evaluations per individual and/or per generation to maximize the advantages while minimizing the number of additional evaluations performed ~\citep{aizawa1994scheduling,branke2003selection,cantu2004adaptive,hansen2008method}. 

As far as we know, the possibility to measure the impact of variations on the fitness measure and on the performance of the evolving robots was not investigated in previous research. Consequently, the measure introduced in this article is original. Moreover,  the impact of the range of variations and the impact of the modality with which variations are introduced were not investigated in previous research.  By modality, we mean the way in which variations are introduced, e.g., whether variations affect the initial state of the agent or the state of the agents during each step and whether the range of variations is constant or varies during the evaluation episode or across generations.

To study these issues we consider the case of neuro-robots, i.e., robots provided with a neural network controller. The architecture of the neural network is fixed. The connection weights are encoded in free parameters and evolved.  Moreover, we study the effect of varying the initial state of the robot and the state of the robot in each step. More specifically, we study the effect of varying the initial posture of the robot and the effects of the motor actions of the robot in each step. We do not consider other forms of variations, e.g., variations affecting the characteristics of the surface over which the robots are located or the position and the characteristics of the objects present in the environment. The impact of other forms of variations should be investigated in further studies. However, we do not see reasons for expecting qualitatively different results with different types of variations. 

Finally, we analyze whether robots evolved in varying conditions outperform robots evolved in non-varying ones independently from whether the evolved robots are evaluated in varying or non-varying conditions. The fact that the introduction of noise in the selection process can lead to the selection of better solutions is well known (see for example ~\citep{branke2003selection}). Here we analyze whether the introduction of morphological variations can produce a functionally similar effect.

For examples of other works addressing the evolution of robots capable to solve the tasks considered in our study see~\citep{meng2022integrating,tjanaka2022scaling,pinosky2022hybrid,kuznetsov2020controlling}

\section{Method}
In this section, we describe the problems, the robots, the controllers of the robots, and the evolutionary algorithm used. Moreover, we introduce a measure that can be used to quantify the impact of variations in the evaluation process. 

\subsection{The Tasks}
\begin{figure}
    \centering
    \includegraphics[width=1.05\textwidth]{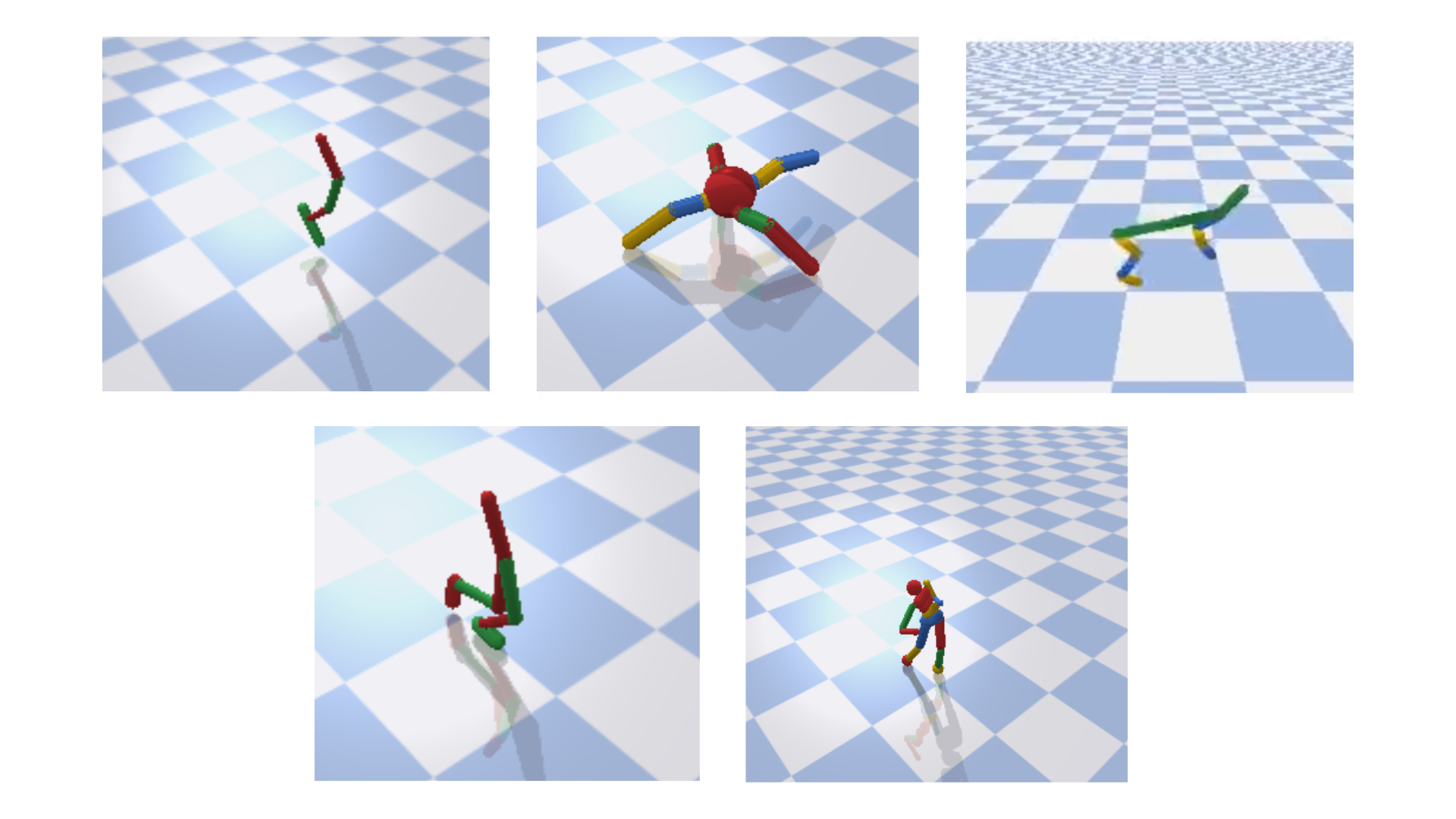}
    \caption{The hopper (top-left), ant (top-center), halfcheetah (top-right), walker2D (bottom-left) and humanoid (bottom-right) robots used in the corresponding Hopper, Ant, Halfcheetah, Walker2d, and Humanoid tasks.}
    \label{fig:robots}
\end{figure}

We will analyze the impact of variations in the case of the Pybullet locomotor tasks ~\citep{coumans2016pybullet} which involve simulated robots located on flat surfaces that should acquire the ability to jump or walk toward a given destination as fast as possible. The Pybullet locomotor tasks are a standard testbed widely used to benchmark evolutionary and reinforcement learning algorithms. More specifically we considered the Hopper, Ant, Halfcheetah, Walker2d, and Humanoid tasks which involve the simulated robots shown in Figure \ref{fig:robots}. We will use the term capitalized to indicate the task (i.e., the robot, the fitness function, and the termination conditions) and the terms without capitalization to indicate the simulated robot used in the task. 

The hopper, halfcheetah, ant, walker2D, and humanoid robots have 3, 6, 8, 6, and 17 actuated hinge joints, respectively. The action vector computed by the robot's neural controller includes $N$ values encoding the torques applied to the N corresponding joints. The observation vector, which constitutes the input of the robot's neural controller, includes the position and orientation of the robot, the angular position and velocity of the joints, and the state of sensors that detect when a body part is in contact with another body. The initial posture of the simulated robot varies randomly within a given range (see below). The evaluation episodes are terminated after 1,000 steps or, prematurely, when the torso of the robots falls below a given threshold. 

The fitness of the robots is computed by summing the rewards obtained during each step of the evaluation episode. The reward functions, which are used to compute the reward gained in each step, include six components: (1) a progress component that encodes the speed of the robot toward the destination, (2) a bonus for staying upright, (3) an electricity cost that corresponds to the average of the dot product of the action vector and of the joint speed vector, (4) a stall cost corresponding to the average of the squared action vector, (5) a cost proportional to the number of joints that reached the corresponding limits, and (6) a cost of -1.0 for falling down. The bonus is set to 2.0 in the case of the Humanoid and to 1.0 in the case of the other tasks. The electricity cost, stall cost, and joint at limit cost are weighted by -8.5, -0.425, and -0.1 in the case of the humanoid, and for -2.0, -0.1, and -0.1 in the case of the other tasks. A summary of the default reward functions can be seen in Table 1. As demonstrated in~\citep{pagliuca2020efficacy}, these reward functions included in the Pybullet library work well for reinforcement learning algorithms such as the PPO ~\citep{Schulman2017} but produce poor results with  evolutionary algorithms. For this reason, we report the results obtained with both the reward functions described above, which work well with reinforcement learning algorithms, and with the rewards functions introduced in ~\citep{pagliuca2020efficacy}, which work well with evolutionary algorithms.  The latter reward functions are simpler and include the first component only in the case of the Hopper and Walker2D tasks, components 1 and 5 in the case of the Halfcheetah task, components 1, 2, 4, and 5, in the case of the Ant task, and for the Humanoid task, same as Ant Task, plus two additional components which penalize by -0.1 the angular offset between the orientation of the robot and the target destination, and a strong penalization of -10.0 for joints over limits. The bonus, component 2, is set to 0.01 and to 0.75 in the case of the Ant and Humanoid tasks. A summary of the reward functions optimized for evolutionary algorithms can be seen in Table 2. We will indicate the experiments performed with the reward functions designed for evolutionary and reinforcement learning algorithms with R-EA and R-RL, respectively. Notice that the experiments reported in this article are all carried on by using the evolutionary algorithm described in section 3.2.

\begin{table}[]
\centering
\caption{Default reward functions (R-RL) components values for each task.}
\begin{tabular}{l|
>{\columncolor[HTML]{9AFF99}}l 
>{\columncolor[HTML]{9AFF99}}c 
>{\columncolor[HTML]{FFCCC9}}c 
>{\columncolor[HTML]{FFCCC9}}c 
>{\columncolor[HTML]{FFCCC9}}c 
>{\columncolor[HTML]{FFCCC9}}c |}
\cline{2-7}
 &
  \multicolumn{6}{c|}{\cellcolor[HTML]{C0C0C0}\textbf{Fitness Components}} \\ \hline
\multicolumn{1}{|l|}{\cellcolor[HTML]{C0C0C0}\textbf{Tasks}} &
  \multicolumn{1}{c|}{\cellcolor[HTML]{9AFF99}\textbf{Progress}} &
  \multicolumn{1}{l|}{\cellcolor[HTML]{9AFF99}\textbf{Upright}} &
  \multicolumn{1}{l|}{\cellcolor[HTML]{FFCCC9}\textbf{Electricity}} &
  \multicolumn{1}{l|}{\cellcolor[HTML]{FFCCC9}\textbf{Stall}} &
  \multicolumn{1}{l|}{\cellcolor[HTML]{FFCCC9}\textbf{\begin{tabular}[c]{@{}l@{}}Joints \\ at Limit\end{tabular}}} &
  \multicolumn{1}{l|}{\cellcolor[HTML]{FFCCC9}\textbf{\begin{tabular}[c]{@{}l@{}}Falling \\ down\end{tabular}}} \\ \hline
\multicolumn{1}{|l|}{Humanoid} &
  \multicolumn{1}{l|}{\cellcolor[HTML]{9AFF99}distance} &
  \multicolumn{1}{c|}{\cellcolor[HTML]{9AFF99}2.0} &
  \multicolumn{1}{c|}{\cellcolor[HTML]{FFCCC9}-8.5} &
  \multicolumn{1}{c|}{\cellcolor[HTML]{FFCCC9}-0.425} &
  \multicolumn{1}{c|}{\cellcolor[HTML]{FFCCC9}-0.1} &
  -1 \\ \hline
\multicolumn{1}{|l|}{\begin{tabular}[c]{@{}l@{}}Hopper, Ant,\\ HalfCheetah,\\ Walker2D  \end{tabular}} &
  \multicolumn{1}{l|}{\cellcolor[HTML]{9AFF99}distance} &
  \multicolumn{1}{c|}{\cellcolor[HTML]{9AFF99}1.0} &
  \multicolumn{1}{c|}{\cellcolor[HTML]{FFCCC9}-2.0} &
  \multicolumn{1}{c|}{\cellcolor[HTML]{FFCCC9}-0.1} &
  \multicolumn{1}{c|}{\cellcolor[HTML]{FFCCC9}-0.1} &
  -1 \\ \hline
\end{tabular}
\end{table}


\begin{table}[]
\centering
\caption{Reward functions components values for each task optimized for evolutionary algorithms (R-EA).}
\begin{tabular}{l|lcccc
>{\columncolor[HTML]{FFCCC9}}c |}
\cline{2-7}
 &
  \multicolumn{6}{c|}{\cellcolor[HTML]{C0C0C0}\textbf{Fitness Components}} \\ \hline
\multicolumn{1}{|l|}{\cellcolor[HTML]{C0C0C0}\textbf{Tasks}} &
  \multicolumn{1}{c|}{\cellcolor[HTML]{9AFF99}\textbf{Progress}} &
  \multicolumn{1}{c|}{\cellcolor[HTML]{9AFF99}\textbf{Upright}} &
  \multicolumn{1}{c|}{\cellcolor[HTML]{FFCCC9}\textbf{Stall}} &
  \multicolumn{1}{c|}{\cellcolor[HTML]{FFCCC9}\textbf{\begin{tabular}[c]{@{}c@{}}Joints \\ at Limit\end{tabular}}} &
  \multicolumn{1}{c|}{\cellcolor[HTML]{FFCCC9}\textbf{\begin{tabular}[c]{@{}c@{}}Joints\\ Over \\ Limit\end{tabular}}} &
  \textbf{\begin{tabular}[c]{@{}c@{}}Angular \\ offset\end{tabular}} \\ \hline
\multicolumn{1}{|l|}{Humanoid} &
  \multicolumn{1}{l|}{\cellcolor[HTML]{9AFF99}distance} &
  \multicolumn{1}{c|}{\cellcolor[HTML]{9AFF99}0.75} &
  \multicolumn{1}{c|}{\cellcolor[HTML]{FFCCC9}-0.01} &
  \multicolumn{1}{c|}{\cellcolor[HTML]{FFCCC9}-0.1} &
  \multicolumn{1}{c|}{\cellcolor[HTML]{FFCCC9}-10.0} &
  -0.1 \\ \hline
\multicolumn{1}{|l|}{Hopper} &
  \multicolumn{1}{l|}{\cellcolor[HTML]{9AFF99}distance} &
  \multicolumn{1}{c|}{\cellcolor[HTML]{9AFF99}0} &
  \multicolumn{1}{c|}{\cellcolor[HTML]{FFCCC9}0} &
  \multicolumn{1}{c|}{\cellcolor[HTML]{FFCCC9}0} &
  \multicolumn{1}{c|}{\cellcolor[HTML]{FFCCC9}0} &
  0 \\ \hline
\multicolumn{1}{|l|}{Walker} &
  \multicolumn{1}{l|}{\cellcolor[HTML]{9AFF99}distance} &
  \multicolumn{1}{c|}{\cellcolor[HTML]{9AFF99}0} &
  \multicolumn{1}{c|}{\cellcolor[HTML]{FFCCC9}0} &
  \multicolumn{1}{c|}{\cellcolor[HTML]{FFCCC9}0} &
  \multicolumn{1}{c|}{\cellcolor[HTML]{FFCCC9}0} &
  0 \\ \hline
\multicolumn{1}{|l|}{HalfCheetah} &
  \multicolumn{1}{l|}{\cellcolor[HTML]{9AFF99}distance} &
  \multicolumn{1}{c|}{\cellcolor[HTML]{9AFF99}0} &
  \multicolumn{1}{c|}{\cellcolor[HTML]{FFCCC9}0} &
  \multicolumn{1}{c|}{\cellcolor[HTML]{FFCCC9}-0.1} &
  \multicolumn{1}{c|}{\cellcolor[HTML]{FFCCC9}0} &
  0 \\ \hline
\multicolumn{1}{|l|}{Ant} &
  \multicolumn{1}{l|}{\cellcolor[HTML]{9AFF99}distance} &
  \multicolumn{1}{c|}{\cellcolor[HTML]{9AFF99}0.01} &
  \multicolumn{1}{c|}{\cellcolor[HTML]{FFCCC9}-0.01} &
  \multicolumn{1}{c|}{\cellcolor[HTML]{FFCCC9}-0.1} &
  \multicolumn{1}{c|}{\cellcolor[HTML]{FFCCC9}0} &
  0 \\ \hline
\end{tabular}
\end{table}

For more details on the tasks and on the reward functions see ~\citep{coumans2016pybullet} and ~\citep{pagliuca2020efficacy}.

\subsubsection{The neural network controller of the robots} The robots are controlled with 3-layered feedforward neural networks, which means an input layer, a hidden layer including 50 internal neurons in the case of the Hopper, Halfcheetah, Ant, and Walker2d problems and 250 internal neurons in the case of the Humanoid problem, and an output layer. The activation of the internal and motor neurons is computed with the hyperbolic tangent (tanh) and linear functions, respectively. The number of input and output neurons varies in different tasks and corresponds to the length of the observation and action vectors of the corresponding robot. The connection weights are encoded in free parameters and evolved with the algorithm described below. 

\subsection{The algorithm}

The robots are evolved by using the OpenAI-ES algorithm~\citep{salimans2017evolution}, i.e., a state-of-the-art algorithm that belongs to the class of natural evolutionary strategies (NES)~\citep{wierstra2014natural,sehnke2010parameter,glasmachers2010exponential,schaul2011high}. Let $F$ denote the objective function acting on parameters $\theta$, the algorithm generates a population with a Gaussian distribution over $\theta$ parameterized by $\psi$ and tries to maximize the mean objective value $\mathbb{E}_{\theta{\sim}p_\psi}F(\theta)$ over the population by using the Adam stochastic optimizer~\citep{wierstra2014natural}. During each iteration, it takes a gradient step by using the estimator presented in equation \ref{saliman-eq}.

\begin{equation} \label{saliman-eq}
\nabla_\psi \mathbb{E}_{\theta{\sim}p_\psi}F(\theta)=\mathbb{E}_{\theta{\sim}p_\psi}\{F(\theta)\nabla_\psi log\ p_\psi(\theta)\}
\end{equation}

The size of the population is set to 500 in the case of the Humanoid task and to 40 in the other tasks. The evolutionary process is continued until the total number of steps exceeds 100 million steps in the case of the Humanoid task and 50 million steps in the case of the other tasks. As in ~\citep{salimans2017evolution} the observation vectors were normalized by using virtual batch normalization ~\citep{salimans2016improved,salimans2017evolution}, the connection weights were normalized by using weight decay~\citep{ng2004feature}, the variance of the perturbations used to generate offspring was set to 0.02, and the step size of the Adam optimizer was set to 0.01. We ran 10 replications for each experimental condition. 

\subsection{Measuring the impact of variations}
Many evolutionary algorithms rely on rank-based selection.  For these algorithms, a straightforward way to measure the impact of variations on the fitness function consists in comparing the ranking of the fitness measures obtained during two independent evaluations. In the absence of variation, the ranking will be identical. Instead, in the presence of variation, the ranking will differ. For example, the individual that resulted the very best in the first evaluation can become the fourteen best in the second evaluation. We can measure the relative impact of variations (IV) by computing the average difference between the ranking positions obtained by each individual in two independent evaluations normalized in the range [0.0, 1.0]:

 \begin{equation} \label{eqn}
	IV = \frac{\sum_{i=1}^{s} \frac{\lvert r1_i - r2_i \rvert}{s-1}  }{s}
\end{equation}

\noindent where $r1_i$ and $r2_i$ are the ranking positions of individual $i$ in the first and in the second independent evaluation, and $s$ is the population size. It is noteworthy that for completely random generated rankings, for large population sizes, this measure follows a Gaussian distribution with an expectation of around 0.333. Therefore, we assume that the signal-to-noise ratio (SNR) is defined by the equation \ref{eqn2} which is defined as follow:

 \begin{equation} \label{eqn2}
	SNR = \frac{0.333 - IV}{0.333}
\end{equation}

We refer to this measure as IV which stands for Impact of Variations. This measure is similar to the one proposed by~\citet{hansen2008method} to vary the number of evaluation episodes allocated to each individual. As mentioned above, varying the number of episodes per individual permits allocating the computational cost required to perform additional evaluations, especially to the individuals who are more affected by variations.

The IV can vary approximately within the range $[0.0, 0.333]$ where $0.0$ corresponds to a situation in which the variations are absent or do not have any effect on the ranking, $0.333$ corresponds to a situation in which the ranking of the individuals reflects exclusively the effect of the variations (i.e., a situation in which the fitness measure does not provide any information about the relative ability of individuals). The value of $0.333$ can be computed experimentally by computing the measure on several sets of rankings generated randomly. Therefore it is the expectation when no fitness signal is present in the rankings. It is important to consider that the IV does not depend only on the range of variations. It also depends on the task and on the behavioral strategy of the robots. 
 
 Overall, it is expected that the greater the range of variations, the greater the value of the IV measure. It is noteworthy though, that the agents can partially neutralize the environmental variation by acting appropriately, i.e., by executing actions that minimize or neutralize the effects of variations. Moreover, as mentioned above, the effect of the variations depends on the task. For example, variations affecting the shape of the objects present in the environment or the observations of the robot are expected to have a small impact on the fitness of a wheeled robot evolved for the ability to move by avoiding obstacles. Instead, the same type of variations is expected to have a significant effect on the fitness of a humanoid robot evolved for the ability to grasp objects. Finally, the IV depends on the behaviors displayed by the population of individuals. The IV measure will result in smaller or higher values depending on whether the behavior of the robots is robust or fragile with respect to morphological variations.

 In the results reported below the IV is computed by evaluating the population twice every generation. 

\section{Experimental Protocol}

In this section, we describe the experiments conducted to study the impact of the variance of variations and of the modalities with which variations are introduced. The variations can affect the state of the robot and/or the environment at the beginning of the evaluation episodes or during each step. Moreover, we describe the experiments conducted to analyze the advantage which can be gained by increasing the precision of the fitness estimation. 

\subsection{Variations affecting the initial state} 
A first type of variation concerns the initial state of the robot and/or of the environment, i.e., the state at the beginning of evaluation episodes. In the case of the tasks considered, the initial state of the robot depends on the position of the joints which connect the parts of the robot's body. In the standard implementation of the Pybullet locomotor tasks, the initial position of the joints is varied with respect to their central position with Gaussian random values which have a variance of 0.1 rad and an average of 0.0 rad. In a first series of experiments, we will thus study the effect of setting the variance of this type of variation to 0.1 rad and 0.03 rad.  We will use our IV measure to study the impact of the amplitude of the variations on the stochasticity of the fitness. Moreover, we will analyze the impact of the distribution of variation on the performance and robustness of the evolved robots. 

\subsection{Variations affecting all states} 
A second type of variation concerns the effect of the motor actions performed in each step. Introducing small variations in the action vector computed by the neural controller of the robot is a common practice that permits taking into account the fact that the effect of physical motors is influenced by factors that are neglected in simulation, e.g., dust, wind, or small irregularities of the terrain. For example, in the experiments reported in~\citep{salimans2017evolution} the action vector is perturbed with Gaussian random values which have a variance of 0.01 and an average of 0.0. To analyze the effect of variations of this kind we will carry out experiments by increasing the variance of variations up to 0.6. 
The amplitude of the variations affecting each state can be constant or can vary during the course of the evaluation episode or across generations. We will investigate the effect of the modality with which variations are introduced by analyzing the following experimental conditions: (i) a fixed condition in which the range of variations is constant and is set to 0.01, 0.3 or 0.6 (ii) an incremental condition, referred as inc-episode, in which the variance of variations increases linearly during the course of the evaluation episode up to 0.36 or 0.55, and (iii) a second incremental condition, referred as inc-generation, in which the variance of variations increases linearly across generations up to 0.55. The rationale behind the inc-episode condition is that it reduces the risk to generate solutions overfitted to a specific range of variation. Moreover, it produces a form of incremental process which exposes low-performing robots, which fall down rapidly, to variations that have a small variance only and  high-performing robots also to variations that have a high variance.  The rationale behind the inc-generation condition is also the generation of an incremental process. Indeed, it exposes the robots of the first generations, which typically display low performance, to variations with a small variance and it exposes the robots of successive generations to variations that have a higher variance.  

The variations described above and in the previous section are applied to all evolving individuals, i.e., to the entire population. 

\subsection{Improving the precision of the fitness}
As mentioned above, the accuracy of the fitness can be improved by using the average value obtained during multiple evaluation episodes in which the robots are exposed to different variations. This method is computationally expensive since evaluation episodes represent the major computational cost. 

To analyze the advantage which can be gained through this method we will report the results of a series of experiments in which the number of evaluation episodes is set to 1, 2, 3, 5, and 10. For the purpose of our comparison, we neglect the computation cost caused by the execution of multiple episodes. This is realized by multiplying the maximum number of evaluation steps by the number of episodes. This implies that the experiments carried out by performing 2 and 10 evaluation episodes, for example, last approximately twice and 10 times more than the experiments carried out by performing a single evaluation episode. 

\subsection{Replicating the experiments}
The experiments can be replicated by using the source code and the instructions available from the following repository: "https://github.com/snolfi/env-variation". 

\section{Results}

In this section, we report the results obtained after conducting the experiments presented in the previous sections. We start by analyzing the impact of the variations affecting the initial state. We analyze the advantages which can be gained by increasing the accuracy of the fitness measure. Finally, we analyze the impact of the variations introduced in each step and the modality with which those variations are introduced. 

\subsection{On the impact of variations affecting the initial state}

\begin{figure*}[htb]
  \centering
  \includegraphics[width=1\linewidth]{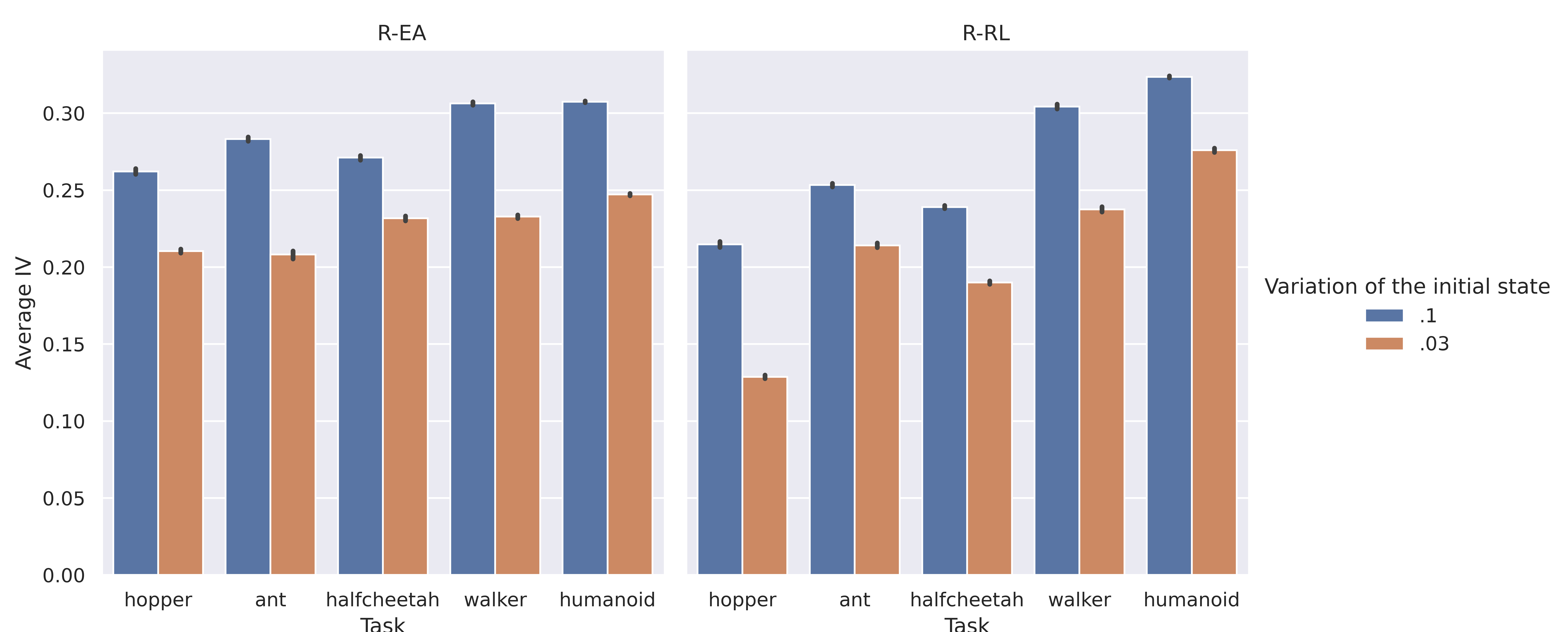}
    \caption{IV for the Hopper, Halfcheetah, Ant, Walker2d and Humanoid tasks. The data shown in blue and orange display the data obtained by using a distribution of 0.1 rad and 0.03 rad to perturb the initial state of the robots, respectively. Values averaged over generations and over 10 replications. Left: experiments performed with the reward function designed for evolutionary algorithms (R-EA). Right: experiments performed with the standard reward function designed for reinforcement learning algorithms (R-RL). }
  \label{fig:iev-bar}
\end{figure*}

\begin{figure*}[htb]
  \centering
  \includegraphics[width=1\linewidth]{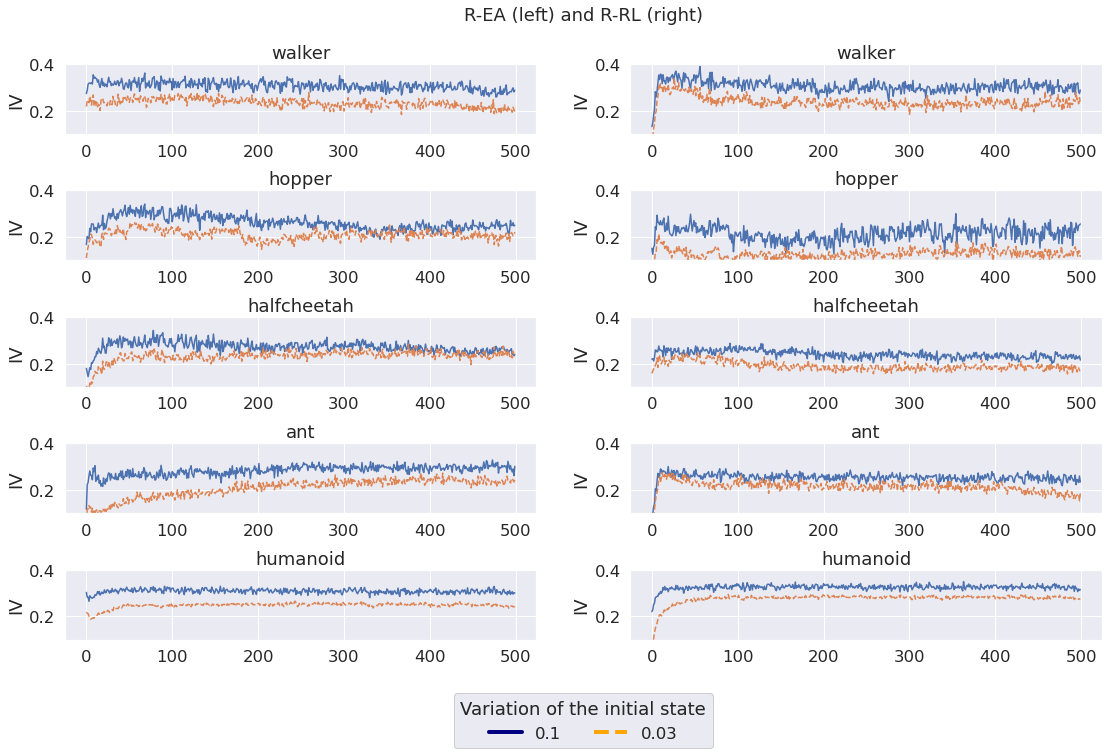}
    \caption{IV across generations. The left and right figures show the data for the experiments performed with the reward function designed for evolutionary (R-EA) and reinforcement learning (R-RL) algorithms. Data averaged over 10 replications. }
  \label{fig:iev-series}
\end{figure*}

Figures \ref{fig:iev-bar} and \ref{fig:iev-series} display the IV and the performance obtained in experiments in which the variance of variations affecting the initial state of the robots is set to 0.1 rad and 0.03 rad, respectively. The variance of variations affecting the action vectors is set to 0.01, i.e., to a small value. The figures on the left and on the right report the results obtained with the reward function designed for evolutionary and reinforcement learning algorithms, respectively. The performance of the evolved agents was computed by post-evaluating them for 10 episodes. 

\begin{figure*}[htb]
  \centering
  \includegraphics[width=1\linewidth]{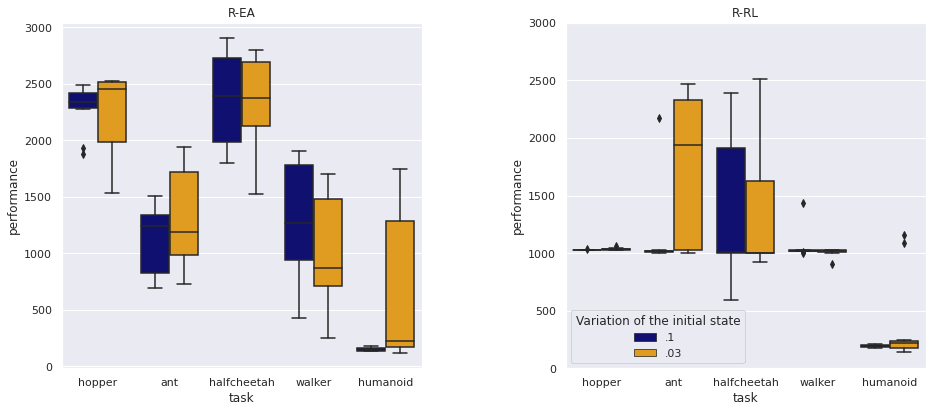}
    \caption{Performance of robots evolved in the experiments in which the variance of the perturbations of the robots' initial state is set to 0.1 rad and 0.03 rad, respectively. The left and right figures show the data for the experiments performed with the reward functions designed for evolutionary (R-EA) and reinforcement learning (R-RL) algorithms. Please notice that the performance measures shown in the left and right figures are not directly comparable since they are computed on the basis of different reward functions. Each boxplot shows the results obtained in 10 replications. Boxes represent the inter-quartile range of the data and horizontal lines inside the boxes mark the median values. The whiskers extend to the most extreme data points within 1.5 times the inter-quartile range from the box.}
  \label{fig:performance}
\end{figure*}

The analysis of the performance and the inspection of the behavior of the agents evolved with the reward function designed for evolutionary algorithms (Figure \ref{fig:performance}, left) show that they display an effective walking behavior in all cases with the exception of the Humanoid task in which the robots fail to walk in the experiments in which the variance of the variations of the initial posture is 0.1 rad. Reducing the amplitude of the variations to 0.03 rad enables the evolving robots to develop effective behaviors also in the case of this task in most of the replications. 

The robots evolved with the reward function designed for reinforcement learning, instead, develop a sub-optimal behavior in most of the replications which consists in quickly assuming a posture from which they can remain still without moving (Figure \ref{fig:performance}, right). Such behavior maximizes the reward bonus received for not falling down but fails to maximize the reward that the robots could potentially receive by jumping or walking forward. In this case the reduction of the distribution of variations to 0.03 rad produces better performance in the case of the Ant, Halfcheetah and Humanoid experiments (Figure \ref{fig:performance}, right, orange data). As we will see in the next section, the convergence of the evolving robots on the suboptimal behavior which consists in staying still depends also on the variance of the variations applied to the action vector in each step. 

The IV is generally quite high and correlates with the complexity of the problem (Figures \ref{fig:iev-bar} and \ref{fig:iev-series}). Indeed, it is higher in the case of the Walker and Humanoid tasks which are more difficult than the other three tasks. The higher difficulty of the former tasks is demonstrated by the fact that performance increases more slowly throughout generations. In the experiment with the fitness function designed for evolutionary algorithms and in which the variance of variation is set to 0.1 rad the signal-to-noise ratio (SNR) ranges from 25\%, in the case of the hopper, to 3\% in the case of the humanoid. 

Overall, these data indicate that the evolutionary algorithm used can tolerate variations which have a rather high impact. In particular, in the experiments performed with the fitness function designed for evolutionary algorithms, the evolutionary process successfully generates effective behavior in all tasks with the exception of the Humanoid task, in which the SNR approaches 0. In the case of this problem, the extremely high value of the IV indicates that the impact of the variations is too high. The problem becomes solvable in the experiments in which the variance of variation of the initial posture is reduced from  0.1 rad to 0.03 rad. 

As expected, the experiments performed with the fitness function designed for reinforcement learning algorithms, which include a bonus for avoiding falling, lead to sub-optimal solutions which achieve a total fitness of approximately 1000 for the first four problems (Figure \ref{fig:performance}). This is explained by the fact that they develop an ability to remain still without walking which enables them to gain a bonus of 1 point for 1000 steps. Interestingly, in the experiments in which the distribution of variation of the initial posture is reduced to 0.03 rad, the agents also manage to develop an ability to walk in the case of the Ant and Halfcheetah problems. In the case of the Humanoid problem, instead, they simply fall down after a few steps.  

\subsection{Improving the quality of the fitness estimation}

\begin{figure*}[htb]
  \centering
  \includegraphics[width=1\linewidth]{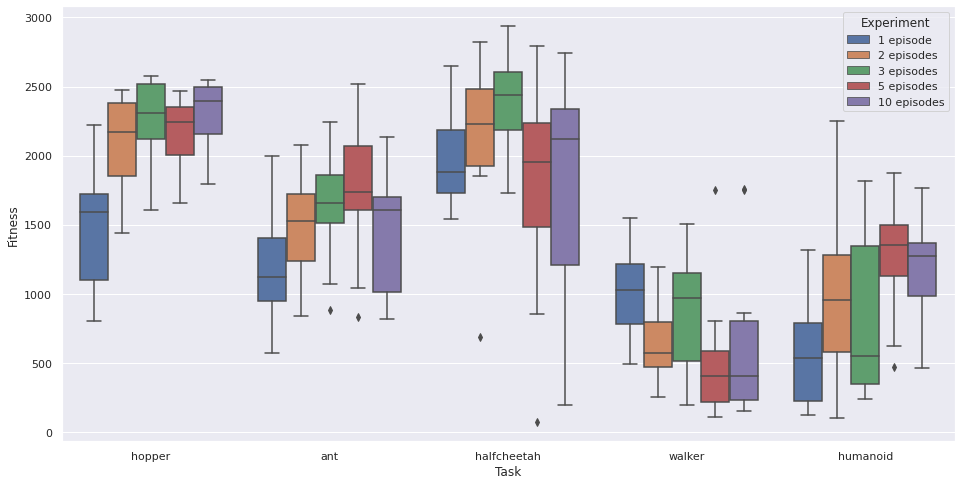}
    \caption{Performance obtained by evaluating the evolving agents for 1, 2, 3, 5, and 10 episodes. Each boxplot shows the results obtained in 10 replications. Boxes represent the inter-quartile range of the data and horizontal lines inside the boxes mark the median values. The whiskers extend to the most extreme data points within 1.5 times the inter-quartile range from the box. Experiments performed with the fitness function designed for the evolutionary algorithm.}
  \label{fig:perf-mult-trials}
\end{figure*}

Figure \ref{fig:perf-mult-trials} display the performance obtained by evaluating the evolving agents for 1, 2, 3, 5, and 10 episodes. The total number of steps performed is multiplied by the number of episodes. This implies that the experiments carried with 2 and 10 episodes last approximately double and 10 times the experiments carried with 1 episode, respectively. The amplitude of the perturbations of the initial posture and of the action vectors is set to 0.1 rad and 0.01, respectively. The performance of the evolved agents is computed by post-evaluating them for 10 episodes.

As expected, the usage of multiple episodes may lead to better performance in certain cases, if one neglects the additional computational cost required to perform multiple episodes. More specifically, the performance is significantly better (Kruskal-Wallis H Test) in the case of the Hopper ($H(5)=17.864, p<.001$), the Ant ($H(5)=23.250, p<.001$) and the Humanoid ($H(5)=25.840, p<.001$) tasks. Surprisingly, instead, performing multiple evaluation episodes does not lead to better performance in the case of the Halfcheetah ($H(5)=5.244, p=0.263$), and produces worse performance in the case of the Walker ($H(5)=25.646, p<.001$).

These results show that reducing the stochasticity of the fitness measure through multiple episodes does not necessarily lead to better performance. This even in problems in which the impact of the environmental variation is high and even ignoring the computation cost required to perform multiple episodes. 

Surprisingly, increasing the accuracy of the fitness measure through multiple episodes can even lead to worse performance. This can be caused by the fact that evaluating the evolving agents for many episodes biases the evolutionary process toward mediocre but reliable strategies, i.e., strategies that are unable to achieve high performance in at least some of the episodes but which minimize the risk to obtain very low performance.

\subsection{On the impact of variations affecting all states}

\begin{figure*}[htb]
  \centering
  \includegraphics[width=1\linewidth]{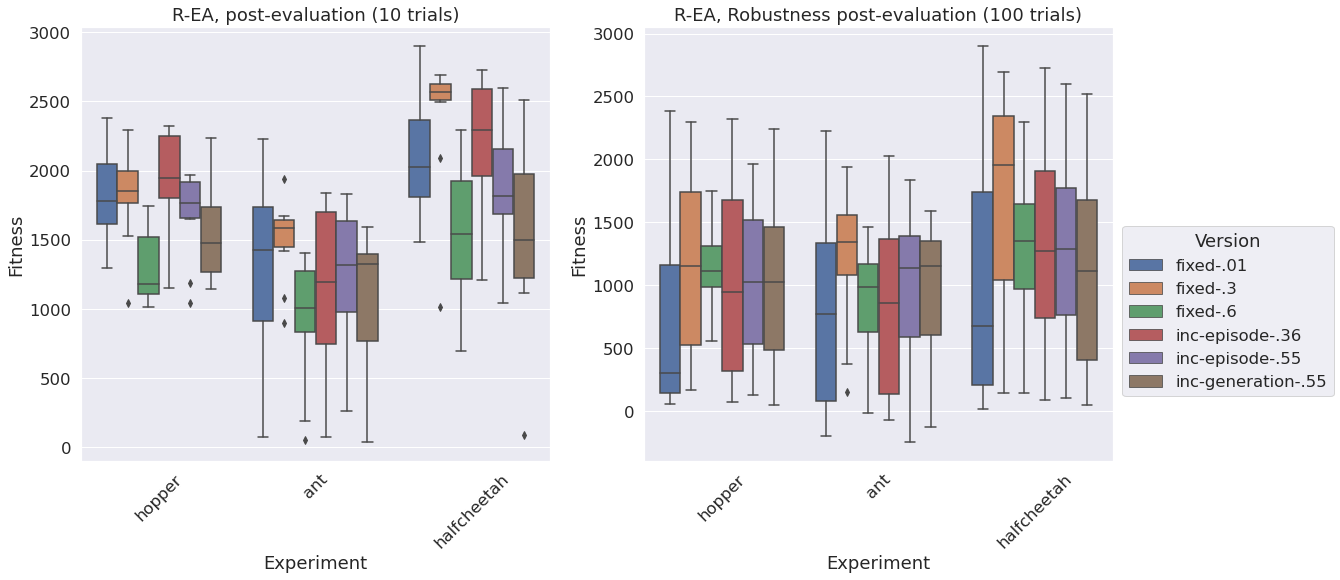}
    \caption{Performance obtained by varying the variance and the modality of action perturbations. Fixed indicates the experiments in which the distribution is constant and is set to 0.01, 0.3, or 0.6), inc-episode the experiments in which the distribution increases linearly during the episode in the range [0.01, 0.36] or [0.01, 0.55], inc-generation the experiments in which the distribution increases linearly across generations in the range [0.01, 0.55]. Experiments performed with the reward function designed for evolutionary algorithms. Each boxplot shows the results obtained in 10 replications. Boxes represent the inter-quartile range of the data and horizontal lines inside the boxes mark the median values. The whiskers extend to the most extreme data points within 1.5 times the inter-quartile range from the box.}
  \label{fig:perf-v5}
\end{figure*}

\begin{figure*}[htb]
  \centering
  \includegraphics[width=1\linewidth]{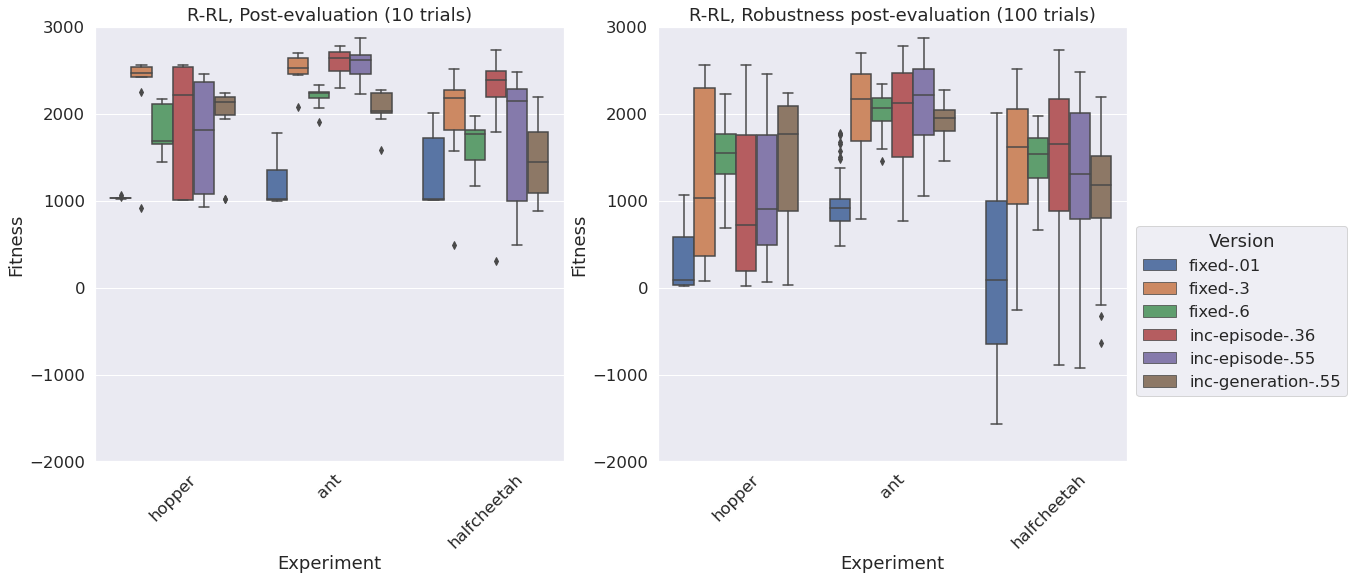}
    \caption{Performance obtained by varying the variance and the modality of action perturbations. Fixed indicates the experiments in which the distribution is constant and is set to 0.01, 0.3 or 0.6), inc-episode the experiments in which the distribution increases linearly during the episode in the range [0.01, 0.36] or [0.01, 0.55], inc-generation the experiments in which the distribution increases linearly across generations in the range [0.01, 0.55]. Experiments performed with the reward function designed for reinforcement learning algorithms. Each boxplot shows the results obtained in 10 replications. Boxes represent the inter-quartile range of the data and horizontal lines inside the boxes mark the median values. The whiskers extend to the most extreme data points within 1.5 times the inter-quartile range from the box.}
  \label{fig:perf-v0}
\end{figure*}

Figure \ref{fig:perf-v5} and \ref{fig:perf-v0} show the performance obtained by varying the variance of variations affecting each state and the modality with which variations are introduced. The former and the latter figures display the results obtained by using the reward function designed for evolutionary and reinforcement learning algorithms, respectively. The variance of the variations affecting the initial state is set to 0.03 rad, i.e., the condition which produced the best results. For this analysis, we restrict our experiments to the Hopper, Halfcheetah, and Ant tasks in which the impact of the variations affecting the initial state is smaller. 

The performance of the evolved agents is computed by post-evaluating them for 10 episodes (left Figures). To evaluate also the robustness of robots we report also the performance obtained in a robustness post-evaluation test in which the evolved robot are post-evaluated for 10 sets of 10 episodes in which the variance of the perturbations affecting the initial state is set to 0.1 rad and the distribution of perturbations affecting the action is set to
[0.01, 0.07, 0.13, 0.19, 0.25, 0.31, 0.37, 0.43, 0.49, 0.55] (right Figures).

As expected, the exposure to large action perturbations permits the development of solutions that are robust with respect to this type of perturbation (Figure \ref{fig:perf-v5} and \ref{fig:perf-v0}, right). 

Interestingly, with the reward function designed for reinforcement learning algorithm, a fixed distribution of 0.3 leads to better results than a fixed distribution of 0.01 even when the robots are post-evaluated in the latter condition (Figure \ref{fig:perf-v0}, left). This is due to the fact that in this set of experiments, the exposure to large action perturbations also reduces the probability that the agents remain trapped in the local minima solutions which consist in standing still without moving. Moreover, also in this case the exposure to large action perturbations permits the development of solutions that are robust with respect to this type of perturbations (Figure \ref{fig:perf-v5}, right). 

The performance differs statistically (Kruskal-Wallis H Test) in all cases ($p<.005$ ) with the exception of the standard post-evaluation results in the case of the ant robots evolved with the reward function designed for evolutionary algorithms ($p=.138$) and of the halfcheetah robots evolved with the reward function designed for reinforcement learning algorithms ($p=.0055$). 

The analysis of the modalities with which variations are introduced did not reveal qualitative differences. More specifically, the incremental modalities in which the variance of variations increases during the evaluation episode or across generations did not lead to better results with respect to the modality in which the variance is constant. Indeed, the best results are obtained in the fixed condition by setting the variance to 0.03 rad. 

\begin{figure*}[htb]
  \centering
  \includegraphics[width=1\linewidth]{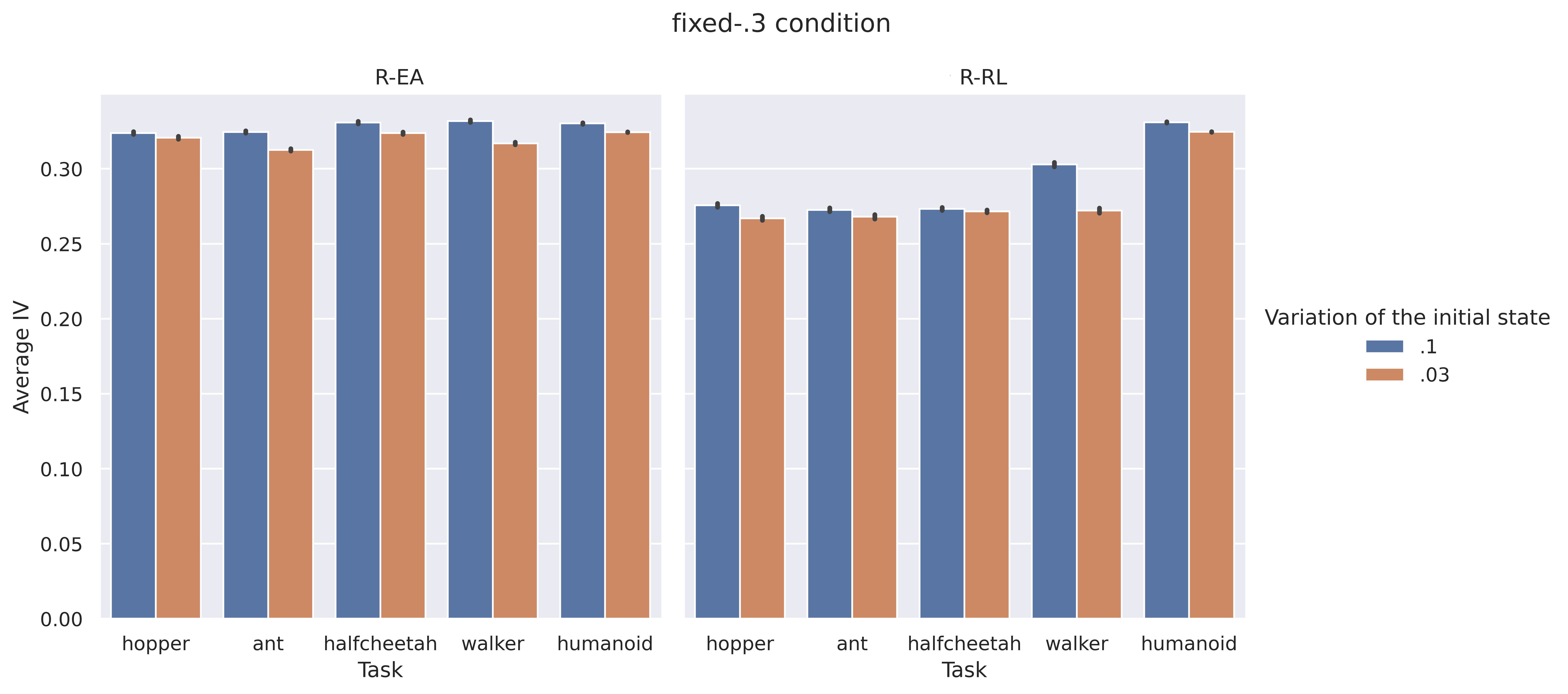}
    \caption{IV for the experiments in which the distribution of the action perturbations is set to 0.3 (fixed-.3 condition). The data visualized in blue and orange correspond to the experiments in which the initial posture of the agents is perturbed with random values with a variance of 0.1 rad and 0.03 rad, respectively. Values averaged over generations and over 10 replications. Left: experiments performed with the reward function designed for evolutionary algorithms (R-EA). Right: experiments performed with the standard reward function designed for reinforcement learning algorithms (R-RL). }
  \label{fig:iev-bar-03}
\end{figure*}

\begin{figure*}[htb]
  \centering
  \includegraphics[width=1\linewidth]{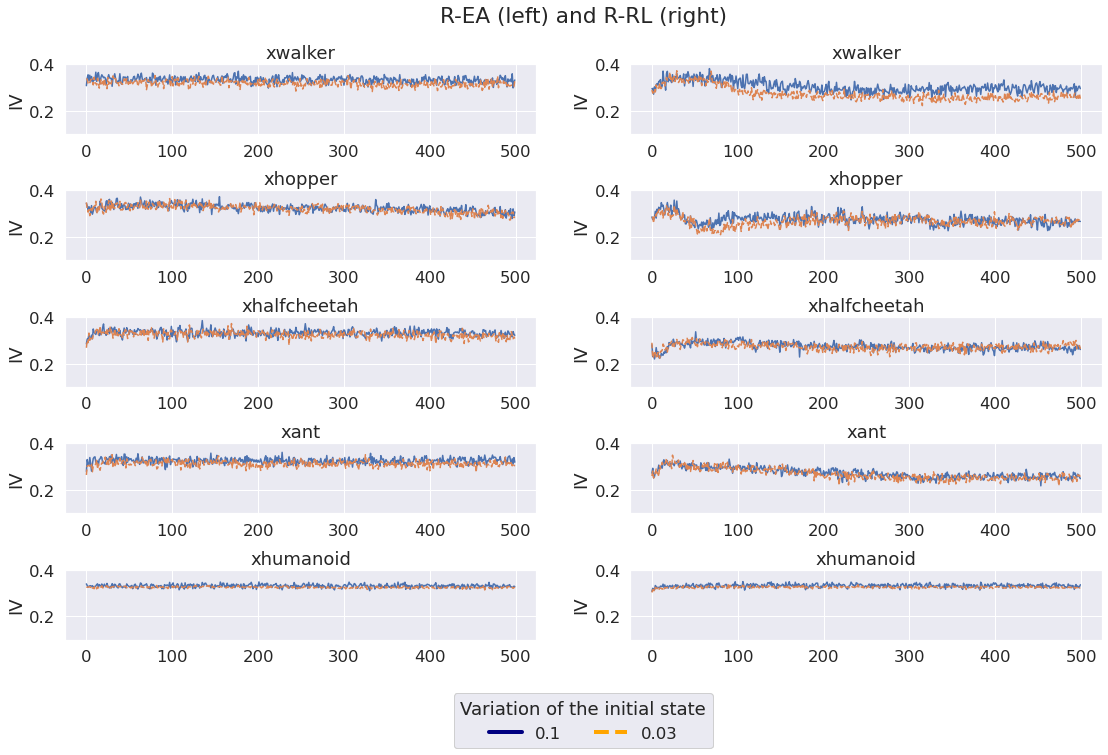}
    \caption{IV values across generations for the experiments in which the variance of action perturbation is set to 0.3 (fixed-.3 condition). Data averaged over 10 replications. Left: experiments performed with the reward functions designed for evolutionary algorithms (R-EA). Right: experiments performed with the reward function designed for reinforcement learning algorithms (R-RL). Data averaged over 10 replications.}
  \label{fig:iev-series-03}
\end{figure*}

Figures \ref{fig:iev-bar-03} and \ref{fig:iev-series-03} show the IV measures in the experimental conditions which lead to the best results overall, i.e., the experiments in which the distribution of the perturbations on the actions is fixed and is set to 0.3 (fixed-.3 condition). As can be seen, the IV measure reaches rather high values. The fact that this condition leads to the best results despite such high IV demonstrates, once more, that the evolutionary algorithm can operate effectively even when the SNR of the fitness measure is extremely low. 

Overall, these results demonstrate that the perturbations introduced in the action vectors during each step are tolerated much better than the perturbations affecting the initial posture of the robots. This is probably explained by the fact that the  impact of the former type of variations unfolds slowly over time and can be reduced by the actions performed by the robots. These variations can thus promote the evolution of solutions which are robust, i.e., are capable to neutralize the effect of variations. The impact of the latter form of variations, instead, affects only the initial part of the evaluation episode.

Moreover, the results described in this section demonstrate that the introduction of large action perturbations enables the evolutionary process to discover solutions that are not only robust to variations but which perform well in general, i.e., which perform well even in the absence of variations. 

\section{Discussion and conclusion}

Exposing evolving robots to variations represents a necessary prerequisite to select individuals capable of operating in natural environments. The lack of variations inevitably leads to the selection of fragile solutions which are fitted to the specific conditions experienced only. To date, however, the determination of the range of variations and of the modalities with which variations are introduced is left to the intuition of the experimenter. These design choices are further complicated by the fact that the impact of variations depends on the problem considered and on the behavioral strategy adopted by the evolving robots. Finally, one should consider that variations do not influence only the robustness of the evolved strategies but also the probability to discover effective solutions.

In this article, we introduced a method for measuring the impact of variations on the fitness measure and we analyzed the relation between the range of variations, the modality with which variations are introduced, and the performance and robustness of the evolved agents. 

The obtained results demonstrate that the evolutionary algorithm can work properly also in the presence of variations which have a rather high impact. This especially in the case of variations which affect the robot in each step, thanks to the fact that the robot can neutralize them by producing the appropriate corrective actions.  Moreover, our results show that improving the precision of the fitness measure through the execution of multiple evaluation episodes does not necessarily provide an advantage, even if one neglects the computation cost required to carry out multiple evaluations.

Overall, these results can be explained by considering that the exposure to varying environmental conditions makes the fitness measure stochastic and by considering that evolutionary algorithms benefit from the usage of stochastic selection processes. Selecting individuals displaying relatively poor performance occasionally improves the exploration capability of the algorithm and reduces the risk to remain trapped in local minima. Indeed, in the case of the experiment reported in this paper, the introduction of significant variations often enables the evolving agents to avoid the local minima constituted by the stay-still strategies that normally emerge in experiments with the reward functions designed for reinforcement learning algorithms. 

These aspects can also explain why robots exposed to large variations can outperform robots exposed to small variations even with respect to the ability to operate in the presence of small variations. The exploration gain induced by variations permits maximizing the chances to discover solutions that are better in both varying and non-varying environments. 

We also showed how the IV measure introduced in this paper can help the experimenter to set the hyperparameters which determine the variance of variations. A high IV value by itself does not necessarily indicate a problem. However, a high IV value observed in experiments in which the evolutionary algorithm fails to produce progress can be used to infer the necessity to reduce the range of variations or to evaluate the agents for multiple episodes.

The analysis of the modalities with which variations are introduced indicates that the variations affecting the state of the agent during each step are tolerated much better than variations affecting the initial posture of the agent. This implies that the latter form of variations should be introduced with more care than the former type of variations.

The analysis of the other modalities with which variations are introduced did not reveal qualitative differences. More specifically, contrary to our expectations, the incremental modalities in which the variance of variations increases during evaluation episodes or across generations did not lead to better results with respect to the modality in which the range of variations is constant.
Overall the analysis reported in this work should enable experimenters to appreciate the importance of variations and to deal with them in a more informed way. 

\small

\bibliographystyle{apalike}
\bibliography{ecjsample}

\begin{thebibliography}{}

\bibitem[Aizawa and Wah, 1994]{aizawa1994scheduling}
Aizawa, A.~N. and Wah, B.~W. (1994).
\newblock Scheduling of genetic algorithms in a noisy environment.
\newblock {\em Evolutionary Computation}, 2(2):97--122.

\bibitem[Arnold and Beyer, 2002]{arnold2002noisy}
Arnold, D.~V. and Beyer, H.-G. (2002).
\newblock {\em Noisy optimization with evolution strategies}, volume~8.
\newblock Springer Science \& Business Media.

\bibitem[Branke, 2012]{branke2012evolutionary}
Branke, J. (2012).
\newblock {\em Evolutionary optimization in dynamic environments}, volume~3.
\newblock Springer Science \& Business Media.

\bibitem[Branke and Schmidt, 2003]{branke2003selection}
Branke, J. and Schmidt, C. (2003).
\newblock Selection in the presence of noise.
\newblock In {\em Genetic and Evolutionary Computation Conference}, pages
  766--777. Springer.

\bibitem[Bredeche et~al., 2012]{bredeche2012environment}
Bredeche, N., Montanier, J.-M., Liu, W., and Winfield, A.~F. (2012).
\newblock Environment-driven distributed evolutionary adaptation in a
  population of autonomous robotic agents.
\newblock {\em Mathematical and Computer Modelling of Dynamical Systems},
  18(1):101--129.

\bibitem[Cant{\'u}-Paz, 2004]{cantu2004adaptive}
Cant{\'u}-Paz, E. (2004).
\newblock Adaptive sampling for noisy problems.
\newblock In {\em Genetic and Evolutionary Computation Conference}, pages
  947--958. Springer.

\bibitem[Coumans and Bai, 2016]{coumans2016pybullet}
Coumans, E. and Bai, Y. (2016).
\newblock Pybullet, a python module for physics simulation for games, robotics
  and machine learning.

\bibitem[Cully et~al., 2015]{cully2015robots}
Cully, A., Clune, J., Tarapore, D., and Mouret, J.-B. (2015).
\newblock Robots that can adapt like animals.
\newblock {\em Nature}, 521(7553):503--507.

\bibitem[Floreano and Urzelai, 2000]{floreano2000evolutionary}
Floreano, D. and Urzelai, J. (2000).
\newblock Evolutionary robots with on-line self-organization and behavioral
  fitness.
\newblock {\em Neural Networks}, 13(4-5):431--443.

\bibitem[Glasmachers et~al., 2010]{glasmachers2010exponential}
Glasmachers, T., Schaul, T., Yi, S., Wierstra, D., and Schmidhuber, J. (2010).
\newblock Exponential natural evolution strategies.
\newblock In {\em Proceedings of the 12th annual conference on Genetic and
  evolutionary computation}, pages 393--400.

\bibitem[Hansen et~al., 2008]{hansen2008method}
Hansen, N., Niederberger, A.~S., Guzzella, L., and Koumoutsakos, P. (2008).
\newblock A method for handling uncertainty in evolutionary optimization with
  an application to feedback control of combustion.
\newblock {\em IEEE Transactions on Evolutionary Computation}, 13(1):180--197.

\bibitem[Harik et~al., 1999]{harik1999gambler}
Harik, G., Cant{\'u}-Paz, E., Goldberg, D.~E., and Miller, B.~L. (1999).
\newblock The gambler's ruin problem, genetic algorithms, and the sizing of
  populations.
\newblock {\em Evolutionary computation}, 7(3):231--253.

\bibitem[Jakobi et~al., 1995]{jakobi1995noise}
Jakobi, N., Husbands, P., and Harvey, I. (1995).
\newblock Noise and the reality gap: The use of simulation in evolutionary
  robotics.
\newblock In {\em European Conference on Artificial Life}, pages 704--720.
  Springer.

\bibitem[Koos et~al., 2012]{koos2012transferability}
Koos, S., Mouret, J.-B., and Doncieux, S. (2012).
\newblock The transferability approach: Crossing the reality gap in
  evolutionary robotics.
\newblock {\em IEEE Transactions on Evolutionary Computation}, 17(1):122--145.

\bibitem[Kuznetsov et~al., 2020]{kuznetsov2020controlling}
Kuznetsov, A., Shvechikov, P., Grishin, A., and Vetrov, D. (2020).
\newblock Controlling overestimation bias with truncated mixture of continuous
  distributional quantile critics.
\newblock In {\em International Conference on Machine Learning}, pages
  5556--5566. PMLR.

\bibitem[Meng et~al., 2022]{meng2022integrating}
Meng, J., Zhu, F., Ge, Y., and Zhao, P. (2022).
\newblock Integrating safety constraints into adversarial training for robust
  deep reinforcement learning.
\newblock {\em Information Sciences}.

\bibitem[Milano et~al., 2019]{milano2019moderate}
Milano, N., Carvalho, J.~T., and Nolfi, S. (2019).
\newblock Moderate environmental variation across generations promotes the
  evolution of robust solutions.
\newblock {\em Artificial life}, 24(4):277--295.

\bibitem[Ng, 2004]{ng2004feature}
Ng, A.~Y. (2004).
\newblock Feature selection, l 1 vs. l 2 regularization, and rotational
  invariance.
\newblock In {\em Proceedings of the twenty-first international conference on
  Machine learning}, page~78.

\bibitem[Nolfi et~al., 2016]{nolfi2016evolutionary}
Nolfi, S., Bongard, J., Husbands, P., and Floreano, D. (2016).
\newblock Evolutionary robotics.
\newblock In {\em Springer handbook of robotics}, pages 2035--2068. Springer.

\bibitem[Pagliuca et~al., 2020]{pagliuca2020efficacy}
Pagliuca, P., Milano, N., and Nolfi, S. (2020).
\newblock Efficacy of modern neuro-evolutionary strategies for continuous
  control optimization.
\newblock {\em Frontiers in Robotics and AI}, 7:98.

\bibitem[Pagliuca and Nolfi, 2019]{pagliuca2019robust}
Pagliuca, P. and Nolfi, S. (2019).
\newblock Robust optimization through neuroevolution.
\newblock {\em PloS one}, 14(3):e0213193.

\bibitem[Pinosky et~al., 2022]{pinosky2022hybrid}
Pinosky, A., Abraham, I., Broad, A., Argall, B., and Murphey, T.~D. (2022).
\newblock Hybrid control for combining model-based and model-free reinforcement
  learning.
\newblock {\em The International Journal of Robotics Research}, page
  02783649221083331.

\bibitem[Risi and Stanley, 2012]{risi2012unified}
Risi, S. and Stanley, K.~O. (2012).
\newblock A unified approach to evolving plasticity and neural geometry.
\newblock In {\em The 2012 International Joint Conference on Neural Networks
  (IJCNN)}, pages 1--8. IEEE.

\bibitem[Sadeghi and Levine, 2016]{sadeghi2016cad2rl}
Sadeghi, F. and Levine, S. (2016).
\newblock Cad2rl: Real single-image flight without a single real image.
\newblock {\em arXiv preprint arXiv:1611.04201}.

\bibitem[Salimans et~al., 2016]{salimans2016improved}
Salimans, T., Goodfellow, I., Zaremba, W., Cheung, V., Radford, A., and Chen,
  X. (2016).
\newblock Improved techniques for training gans.
\newblock {\em Advances in neural information processing systems}, 29.

\bibitem[Salimans et~al., 2017]{salimans2017evolution}
Salimans, T., Ho, J., Chen, X., Sidor, S., and Sutskever, I. (2017).
\newblock Evolution strategies as a scalable alternative to reinforcement
  learning.
\newblock {\em arXiv preprint arXiv:1703.03864}.

\bibitem[Salvato et~al., 2021]{salvato2021}
Salvato, E., Fenu, G., Medvet, E., and Pellegrino, F.~A. (2021).
\newblock Crossing the reality gap: a survey on sim-to-real transferability of
  robot controllers in reinforcement learning.
\newblock {\em IEEE Access}.

\bibitem[Schaul et~al., 2011]{schaul2011high}
Schaul, T., Glasmachers, T., and Schmidhuber, J. (2011).
\newblock High dimensions and heavy tails for natural evolution strategies.
\newblock In {\em Proceedings of the 13th annual conference on Genetic and
  evolutionary computation}, pages 845--852.

\bibitem[Schulman et~al., 2017]{Schulman2017}
Schulman, J., Wolski, F., Dhariwal, P., Radford, A., and Klimov, O. (2017).
\newblock Proximal policy optimization algorithms.
\newblock {\em arXiv preprint arXiv:1707.06347}.

\bibitem[Sehnke et~al., 2010]{sehnke2010parameter}
Sehnke, F., Osendorfer, C., R{\"u}ckstie{\ss}, T., Graves, A., Peters, J., and
  Schmidhuber, J. (2010).
\newblock Parameter-exploring policy gradients.
\newblock {\em Neural Networks}, 23(4):551--559.

\bibitem[Stagge, 1998]{stagge1998averaging}
Stagge, P. (1998).
\newblock Averaging efficiently in the presence of noise.
\newblock In {\em International Conference on Parallel Problem Solving from
  Nature}, pages 188--197. Springer.

\bibitem[Tjanaka et~al., 2022]{tjanaka2022scaling}
Tjanaka, B., Fontaine, M.~C., Kalkar, A., and Nikolaidis, S. (2022).
\newblock Scaling covariance matrix adaptation map-annealing to
  high-dimensional controllers.
\newblock In {\em Deep Reinforcement Learning Workshop NeurIPS 2022}.

\bibitem[Wierstra et~al., 2014]{wierstra2014natural}
Wierstra, D., Schaul, T., Glasmachers, T., Sun, Y., Peters, J., and
  Schmidhuber, J. (2014).
\newblock Natural evolution strategies.
\newblock {\em The Journal of Machine Learning Research}, 15(1):949--980.

\end{thebibliography}

\end{document}